\documentclass[runningheads]{llncs}

\usepackage{eccv}

\usepackage{eccvabbrv}

\usepackage{graphicx}
\usepackage{booktabs}

\usepackage[accsupp]{axessibility}  

\usepackage{hyperref}

\usepackage{orcidlink}

\usepackage{tabularx}
\usepackage{makecell} 
\usepackage[table]{xcolor}
\usepackage{booktabs}

\definecolor{bestc}{HTML}{FFA8B5}    
\definecolor{secondc}{HTML}{FFCBA4}  
\definecolor{thirdc}{HTML}{FFE894}   

\newcommand{\best}[1]{\cellcolor{bestc}\textbf{#1}}
\newcommand{\second}[1]{\cellcolor{secondc}#1}
\newcommand{\third}[1]{\cellcolor{thirdc}#1}
\usepackage{colortbl}

\newcommand{\capbest}[1]{\colorbox{bestc}{\textbf{#1}}}
\newcommand{\capsecond}[1]{\colorbox{secondc}{#1}}
\newcommand{\capthird}[1]{\colorbox{thirdc}{#1}}

\newcolumntype{Y}{>{\centering\arraybackslash}X}
\usepackage{multirow}
\usepackage{array}

\begin{document}

\title{ACE-GS: Acing the Trade-off with Accurate, Compact and Efficient 3D Gaussian Splatting}

\titlerunning{ACE-GS}

\author{Jijian Zhao\inst{1}}

\authorrunning{J.~Zhao}

\institute{
Huazhong University of Science and Technology, Wuhan, China\\
\email{jijianzhao28@gmail.com}
}

\maketitle

\begin{figure}[h]
  \centering
  \includegraphics[width=\textwidth]{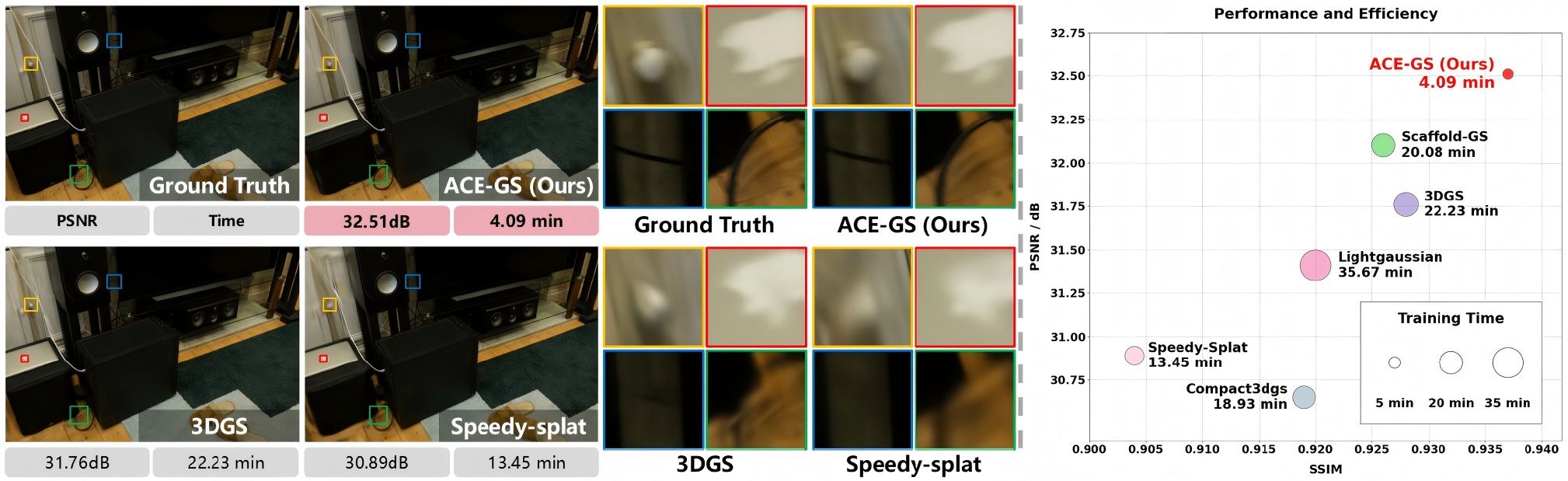}
  \caption{ACE-GS fundamentally breaks the trade-off between optimization speed and rendering fidelity. As demonstrated in the Room scene from the Mip-NeRF 360 dataset~\cite{barron2022mip}, while Speedy-Splat~\cite{hanson2025speedy} suffers from severe blurring, our method successfully recovers sharp high-frequency details such as the Apple logo and thin wires on the floor. Quantitatively, our framework achieves a highly superior balance across the performance and efficiency landscape. Notably, requiring a mere 4 minutes of training, our method not only secures the highest structural similarity but also significantly outperforms the original 3DGS~\cite{kerbl20233d} by 0.75 dB in PSNR.}
  \label{fig:teaser}
\end{figure}

\begin{abstract}
3D Gaussian Splatting achieves exceptional real-time rendering, but its substantial computational and storage demands hinder widespread deployment. Existing accelerated paradigms often aggressively prune primitives for rapid convergence, causing severe loss of high-frequency details. To address this, we tackle the fundamental problem of achieving both exceptional rendering quality and ultra-fast reconstruction speed. In this paper, we propose ACE-GS, a progressive optimization framework tailored for accurate, compressed, and efficient scene representation. We realize that precise primitive management is the key to breaking this trade-off. Therefore, we first design a momentum consistency-guided densification strategy, strictly constraining primitive growth onto authentic geometric manifolds to avoid computational waste while significantly accelerating convergence. Building upon this efficient initialization, we deploy a statistical sensitivity-driven sparsification mechanism to precisely prune redundant primitives, yielding a further compressed footprint. Finally, to thoroughly compensate for the risk of micro-structure loss caused by the aforementioned strict primitive control, we introduce a cross-dimensional residual frequency compensation scheme that explicitly back-injects high-frequency error energy into primitive attributes, perfectly restoring sharp geometric details. Extensive experiments validate our superiority. While maintaining a highly compact scene representation, our system achieves up to 3.7 times training acceleration against the rapid framework Speedy-Splat. Requiring only 3 to 5 minutes to converge, ACE-GS secures the highest structural similarity and achieves a peak PSNR improvement of up to 0.89 dB over the original 3DGS, establishing a new benchmark for ultra-fast and high-fidelity novel view synthesis.
\end{abstract}

\section{Introduction}

Novel view synthesis has long been a central topic in computer vision and graphics \cite{bao20253d, fei20243d}. Recently, 3D Gaussian Splatting has revolutionized the field by introducing an explicit scene representation coupled with an efficient tile-based rasterization pipeline, achieving exceptional real-time rendering performance while maintaining high reconstruction quality \cite{chen2024survey}. Compared with implicit neural representation methods \cite{mildenhall2021nerf} that rely on heavy multilayer perceptron sampling along rays, 3DGS optimizes millions of anisotropic Gaussian primitives to fit complex scene geometries and appearances. This paradigm not only guarantees superior rendering fidelity but also demonstrates highly competitive rendering speeds, making it an important choice for real-time applications such as virtual reality and digital twins \cite{luiten2024dynamic, wu20244d}. The combination of high-quality rendering and differentiable pipelines has led to the widespread adoption of 3DGS, establishing a new benchmark for three-dimensional reconstruction \cite{zhang2024fregs, yu2024mip, ren2024octree}.

\begin{figure}[t]
    \centering
    \includegraphics[width=\linewidth]{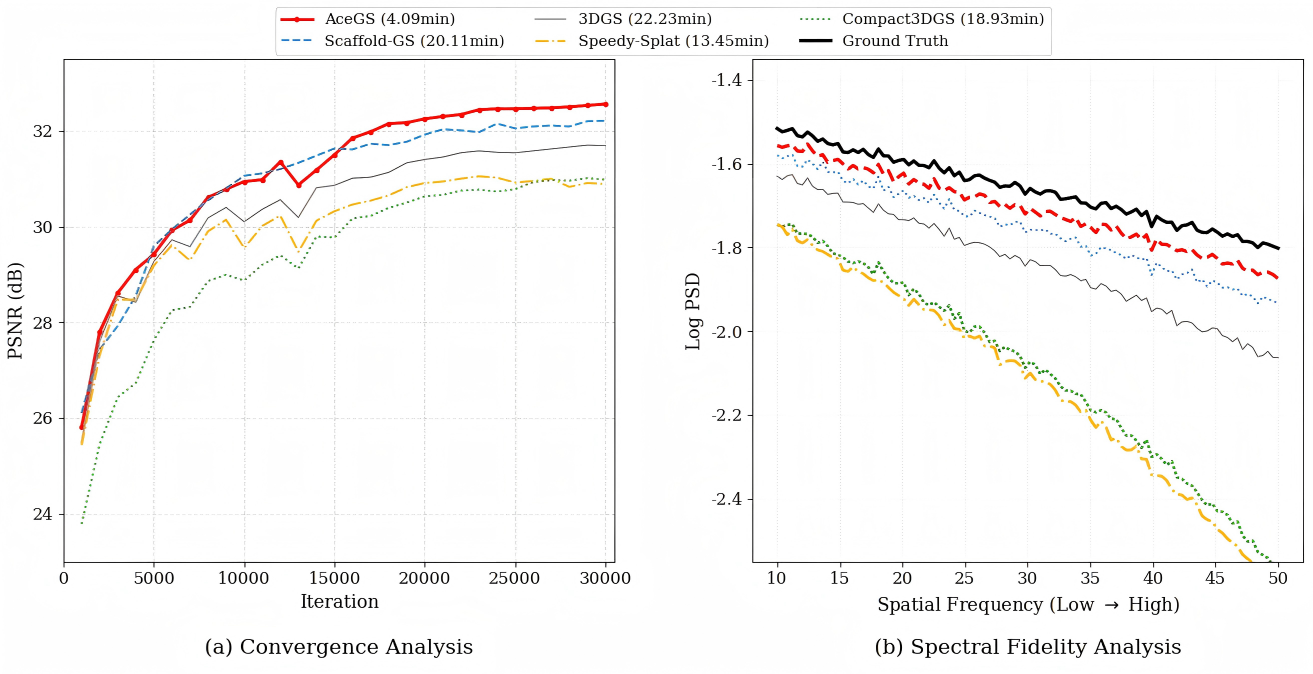}
    \caption{\textbf{Convergence and Spectral Fidelity Analysis on the Room Scene.} (a) PSNR convergence comparison. Our ACE-GS achieves a superior reconstruction quality of 32.51 dB, establishing a significant accuracy lead over both standard and rapid baselines. (b) Spectral analysis of the rendered images. ACE-GS maintains higher energy density in high-frequency bands, aligning closely with the Ground Truth and effectively restoring intricate textures that are typically lost in other accelerated frameworks.}
    \label{fig:motivation}
\end{figure}

However, existing 3DGS frameworks face an irreconcilable contradiction when pursuing dual improvements in reconstruction accuracy and operational efficiency. Although current accelerated paradigms \cite{hanson2025speedy, hanson2025pup} strive to exchange rapid convergence for aggressive primitive pruning, such performance gains often come at the expense of visual fidelity. This crude reduction of primitives to pursue speed leads to the severe loss of critical high-frequency details, see Fig.~\ref{fig:motivation}b. Over-simplified scene representations \cite{szymanowicz2024splatter} not only fail to capture complex geometric structures at the micro-level but also cause significant over-smoothing in the rendering results. In essence, the current acceleration paradigm remains trapped in a trade-off dilemma, failing to achieve a true win-win situation for both ultra-fast reconstruction and high-fidelity output.

To break this trade-off, we conducted a deep analysis of the technical characteristics of Gaussian primitive evolution during the training process. We found that the key to achieving both accuracy and efficiency lies in the precise and dynamic management of primitives. An intuitive concept is that Gaussian primitives in a scene exhibit distinct qualities. By preserving high-quality Gaussians that depict critical structures while eliminating redundant or noisy low-quality ones, we can maintain extreme reconstruction quality while minimizing computational costs \cite{fan2024lightgaussian, wang2024end}. Therefore, a fundamental question remains regarding how to accurately remove low-quality Gaussians while preserving high-quality ones to ensure a compact yet highly expressive scene representation \cite{lee2024compact, fang2024mini, chen2024hac}. To this end, we attempt to tackle the fundamental problem of achieving both exceptional rendering effects and ultra-fast reconstruction speeds.

We propose ACE-GS, a framework designed for Accurate, Compact, and Efficient 3D Gaussian reconstruction. Its core concept revolves around a synergistic scheduling strategy: leveraging momentum consistency-guided densification for rapid convergence, employing statistical sensitivity-driven pruning to eliminate spatial redundancy, and introducing residual frequency compensation for high-fidelity detail recovery. By prioritizing computational resources for core geometric and texture manifolds, ACE-GS achieves significant acceleration while establishing a new benchmark for rendering quality. Our main contributions are summarized as follows:

\begin{itemize}
    \item[$\bullet$] \textbf{Unified and efficient optimization architecture.} We establish a progressive framework that breaks the deadlock between accuracy and efficiency by introducing a rigorous primitive evaluation strategy that ensures only truly effective and high-quality primitives participate in the final scene construction.
    \item[$\bullet$] \textbf{Synergistic primitive optimization and refinement strategy.} Within this framework, we introduce momentum consistency-guided densification for significantly accelerating convergence, employ statistical sensitivity-driven sparsification for maintaining a highly compact scene representation, and propose a cross-dimensional residual frequency compensation scheme for substantially enhancing reconstruction accuracy.
    \item[$\bullet$] \textbf{SOTA performance and exceptional acceleration.} Extensive experiments on 3 challenging datasets demonstrate that ACE-GS achieves state-of-the-art results. Compared with the rapid framework Speedy-Splat, our system achieves up to 2.78 / 3.70 / 3.90 times training acceleration while securing superior performance across all key metrics including PSNR, SSIM, and LPIPS.
\end{itemize}

\section{Related Work}

\indent \textbf{Accurate Rendering and Frequency Modeling.} Novel view synthesis evolves by balancing rendering speed and visual fidelity. While 3D Gaussian Splatting achieves real-time performance, it suffers from aliasing artifacts due to sampling frequency mismatch. Mip-Splatting \cite{yu2024mip} introduces 3D smoothing filters to limit frequencies below the Nyquist limit, while StopThePop \cite{radl2024stopthepop} eliminates multi-resolution flickering via depth-sorted clustering. To improve geometric fidelity, 2DGS \cite{huang20242d} and GOF \cite{yu2024gaussian} constrain Gaussians to explicit surfaces for sharper edges. For appearance modeling, Spec-Gaussian \cite{specgaussian} and GaussianPro \cite{cheng2024gaussianpro} enhance the representation of view-dependent specular highlights \cite{jiang2024gaussianshader}, while PhysGaussian \cite{xie2024physgaussian} incorporates physical laws to model complex dynamics. Additionally, FreGS \cite{zhang2024fregs} and Octree-GS \cite{ren2024octree} further facilitate high-frequency detail recovery by introducing frequency-domain supervision and hierarchical levels of detail \cite{liang2024analytic}. However, these high-precision solutions often rely on heavy regularization or auxiliary structures, increasing memory and optimization costs. They remain less suitable for rapid training. In contrast, we propose a cross-dimensional residual frequency compensation mechanism. By mapping spatial residuals to the frequency domain, we extract high-frequency error energy and back-inject it into primitive attributes, achieving significant detail restoration without additional training burdens.

\indent \textbf{Compact Representation and Online Sparsification.} The massive memory footprint of 3DGS necessitates compact representations for efficient deployment. Research generally follows two paths. Value-based compression methods, such as LightGaussian \cite{fan2024lightgaussian} and Mini-Splatting \cite{fang2024mini}, prune primitives via importance scores, while Compact3DGS \cite{lee2024compact}, Eagles \cite{girish2024eagles}, and Compressed 3DGS \cite{niedermayr2024compressed} adopt quantization to reduce attribute bit-rates. Structural approaches, such as HAC \cite{chen2024hac} and MesonGS \cite{xie2024mesongs}, utilize hash-grids and entropy minimization to eliminate spatial redundancy, while Reduced-3DGS \cite{papantonakis2024reducing} and P-3DGS \cite{hanson2025pup} perform fine-grained selective compression. Although these methods reduce model size, most rely on post-processing stages like distillation or multi-stage fine-tuning. This contradicts the goal of instant reconstruction. Furthermore, static pruning criteria fail to adapt to dynamic optimization. We instead propose a statistical sensitivity-driven online sparsification mechanism. This approach evaluates the expected contribution of each primitive to the global photometric loss directly within the training loop. By employing dual-threshold filtering, we remove redundant primitives in real-time, constructing a lean yet expressive scene backbone without expensive post-processing.

\indent \textbf{Efficient Optimization and Motion Guidance.} Achieving minute-level reconstruction requires significant exploration of optimization efficiency. VastGaussian \cite{lin2024vastgaussian}, CityGaussian \cite{liu2024citygaussian}, and Dogs \cite{chen2024dogs} adopt divide-and-conquer strategies for large-scale scenes . Feed-forward networks like PixelSplat \cite{charatan2024pixelsplat} and MVSplat \cite{chen2024mvsplat} predict parameters directly from multi-view images but often lack generalization. Optimization-centric methods like Gaussianpro \cite{cheng2024gaussianpro} incorporate priors to guide faster convergence \cite{chen2024gaussianeditor}. Recently, rapid frameworks such as Speedy-Splat \cite{hanson2025speedy}, Ges \cite{hamdi2024ges}, and Sugar \cite{guedon2024sugar} achieved impressive speeds through simplified logic. However, these rely on heuristic pruning that lacks adaptive perception of complex geometry, risking the loss of critical micro-structures. We improve upon this by introducing an inherent momentum consistency-guided densification strategy. By tracking the kinematic phase-space trajectories of primitives, we adaptively separate unstructured noise from manifold-aligned points. This strategy constrains primitive growth to well-defined geometric manifolds, ensuring rapid convergence without compromising the reconstruction quality of high-frequency structures.

\section{Preliminary}
\label{sec:preliminary}

\indent 3D Gaussian Splatting~\cite{kerbl20233d} models the 3D scene as a set of anisotropic Gaussians $\mathcal{G} = \{ \mu_k, \Sigma_k, \alpha_k, c_k \}_{k=1}^N$. Each primitive is parameterized by a mean position $\mu \in \mathbb{R}^3$, opacity $\alpha \in [0, 1]$, and spherical harmonics coefficients $c$ for view-dependent color. To ensure positive semi-definiteness during optimization, the 3D covariance matrix $\Sigma$ is defined by a scaling matrix $S$ derived from a vector $s \in \mathbb{R}^3$ and a rotation matrix $R$ constructed from a quaternion $q \in \mathbb{R}^4$:
\begin{equation}
    \Sigma = R S S^T R^T.
    \label{eq:covariance}
\end{equation}

\indent To render the scene from a specific viewpoint, 3D Gaussians are projected onto the 2D image plane. Given the viewing transformation $W$, the covariance matrix $\Sigma'$ in camera coordinates is computed as:
\begin{equation}
    \Sigma' = J W \Sigma W^T J^T,
    \label{eq:projection}
\end{equation}
where $J$ denotes the Jacobian of the affine approximation of the projective transformation. For a given pixel, the final color $C$ is obtained by sorting the overlapping Gaussians $\mathcal{N}$ by depth and accumulating their contributions via $\alpha$-blending:
\begin{equation}
    C = \sum_{i \in \mathcal{N}} c_i \alpha_i \prod_{j=1}^{i-1} (1 - \alpha_j),
    \label{eq:rendering}
\end{equation}
where $\alpha_i$ represents the effective opacity evaluated by the 2D Gaussian distribution. The entire representation is optimized end-to-end by minimizing the photometric loss between rendered images and ground truth views.


\section{Methodology}

\begin{figure}[t]
    \centering
    \includegraphics[width=\textwidth]{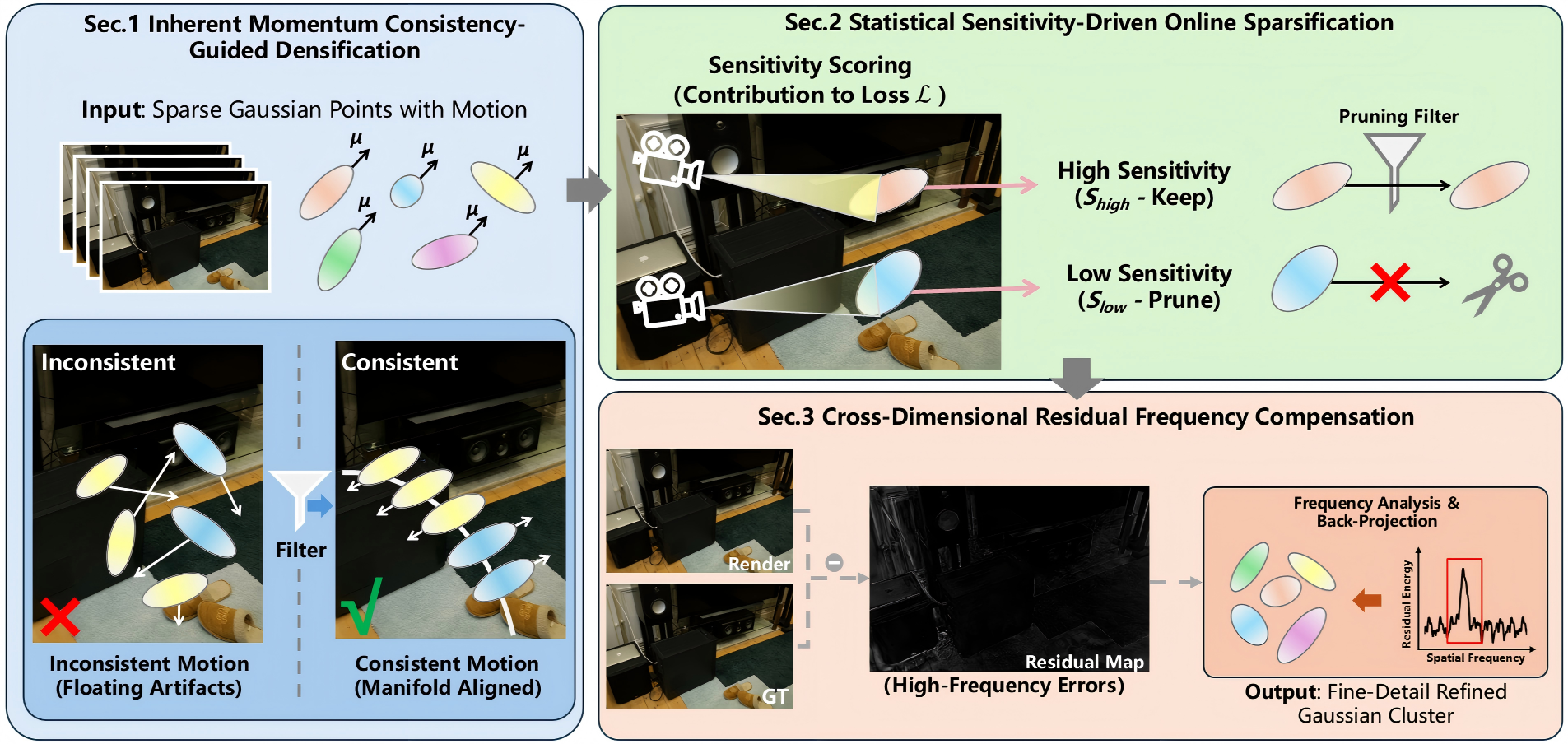}
    \caption{\textbf{Architecture of the ACE-GS synergistic framework.} Our system redefines 3D reconstruction as a dynamic primitive allocation process for fast and high-precision reconstruction. The pipeline utilizes an efficient engine for momentum-guided densification to accelerate convergence and suppress artifacts, a compact engine for sensitivity-driven sparsification to eliminate redundancies, and an accurate engine for frequency compensation to refine details. This mechanism ensures that primitives are precisely allocated to regions with rich visual details, establishing an optimal trade-off between reconstruction efficiency and high rendering fidelity.}
    \label{fig:pipeline}
\end{figure}

\subsection{Overall Architecture and Synergistic Primitive Allocation}
The core design of ACE-GS lies in redefining 3D scene reconstruction as a dynamic and precise primitive allocation process, aiming to overcome the limitations of slow convergence and excessive redundancy in conventional methods when dealing with complex geometries. As illustrated in Fig.~\ref{fig:pipeline}, the framework is not a simple stack of isolated modules but a synergistic closed-loop system. First, the system captures the kinematic characteristics of primitives through the geometric construction module shown in Sec. 1, establishing a robust geometric backbone in the early stages. Subsequently, the compact optimization strategy depicted in Sec. 2 intervenes during the optimization loop to precisely eliminate ineffective redundancies by monitoring each primitive's dynamic contribution to the global loss. Finally, the accuracy enhancement module in Sec. 3 models rendering residuals in the frequency domain, refining high-frequency textures that are difficult to fit in the spatial domain through energy back-injection. This synergistic allocation mechanism ensures that computational resources are consistently concentrated on regions with rich visual details or high geometric complexity, establishing an optimal trade-off between reconstruction efficiency and high rendering fidelity.

\subsection{Inherent Momentum Consistency-Guided Densification}
To overcome the limitations of standard gradient-based strategies in creating floating artifacts in non-manifold regions, we propose a densification scheme guided by kinematic phase-space trajectories. Driven by the motivation that historical motion trends are more deterministic than single-step gradient snapshots, we introduce momentum $m_t$ as a latent variable, defined by the exponential moving average of gradients:
\begin{equation}
    m_t = \gamma \cdot m_{t-1} + (1 - \gamma) \cdot \nabla_p \mathcal{L}_t
\end{equation}
In this equation, $\gamma$ represents the momentum decay factor that balances historical trends and current observations, while $\nabla_p \mathcal{L}_t$ denotes the gradient of the loss function $\mathcal{L}$ with respect to the primitive attribute set $p = \{x, \alpha, s, r, c\}$ at the current iteration $t$. Specifically, we quantify the growth stability through a momentum consistency score $\Psi_i$, which measures the cosine similarity between the current gradient vector and the accumulated momentum:
\begin{equation}
    \Psi_i = \frac{m_t \cdot \nabla_p \mathcal{L}_t}{\|m_t\| \cdot \|\nabla_p \mathcal{L}_t\| + \epsilon}
\end{equation}
where $\epsilon$ is a numerical stability constant introduced to prevent division by zero. To achieve scene-adaptive manifold identification without manual hyperparameter search, we define a dynamic densification trigger function $\mathbb{I}_{densify}$ that compares individual consistency against the mean consistency level $\bar{\Psi}$ of the current primitive population:
\begin{equation}
    \mathbb{I}_{densify} = (\|\nabla \mathcal{L}\| > \tau_{grad}) \land (\Psi_i > \bar{\Psi})
\end{equation}
In this decision logic, $\tau_{grad}$ denotes the predefined gradient magnitude threshold. Consequently, our framework only triggers densification when a primitive's consistency is significantly higher than the current population average, effectively suppressing unstructured noise.

\subsection{Statistical Sensitivity-Driven Online Sparsification}
To maintain model compactness without the prohibitive time cost of offline pruning, we incorporate an online sparsification mechanism that evaluates the actual contribution of each primitive within the optimization loop. We define the statistical sensitivity $S_i$ as the expected gradient magnitude of the global photometric loss over a fixed accumulation window:
\begin{equation}
    S_i = \mathbb{E} \left[ \left| \frac{\partial \mathcal{L}}{\partial \alpha_i} \right| \right]
\end{equation}
In this expression, $\partial \mathcal{L} / \partial \alpha_i$ signifies the partial derivative of the loss function with respect to the opacity $\alpha$ of the $i$-th primitive. These scores are subsequently passed through a pruning filter where we execute online sparsification by defining the primitive survival state $\mathcal{S}_i$ with the following logical expression:
\begin{equation}
    \mathcal{S}_i = \begin{cases} \text{keep} & S_i \geq \tau_{low} \\ \text{prune} & S_i < \tau_{low} \end{cases}
\end{equation}
Here, $\tau_{low}$ represents the sparsification lower-bound threshold that controls the compactness of the model. This sensitivity-driven approach ensures that the model remains refined throughout the training process.

\subsection{Cross-Dimensional Residual Frequency Compensation}
Aiming at the inherent blurring artifacts of Gaussian kernels, we introduce a frequency-domain compensation scheme that targets high-frequency details difficult to fit in the spatial domain. We analyze the spatial residual map $R$, which is the difference between the ground truth image $I_{gt}$ and the current rendered image $I_{render}$, and transform it into the frequency domain via a Fast Fourier Transform $\mathcal{F}$ to extract high-frequency components $\hat{R}_{high}$:
\begin{equation}
    \hat{R}_{high} = \text{HighPass}(\mathcal{F}(I_{gt} - I_{render}))
\end{equation}
In this formulation, $\text{HighPass}(\cdot)$ denotes the high-pass filtering operation used to isolate fine texture details. Subsequently, we project this energy back into the primitive attributes through an inverse mapping. The update amount $\Delta p_i$ is calculated by the following formula:
\begin{equation}
    \Delta p_i = \mathcal{F}^{-1}(\lambda \cdot \hat{R}_{high} \cdot W_i)
\end{equation}
Where $\mathcal{F}^{-1}$ represents the inverse Fast Fourier Transform, $\lambda$ denotes the scaling coefficient that controls the intensity of the frequency injection, and $W_i$ is a normalized spatial weighting term used to precisely distribute frequency energy to the corresponding spatial locations. This frequency-domain feedback provides a more direct path for recovering details.

\section{Experiments}
\label{sec:experiments}

\subsection{Experimental Settings}

\begin{table}[t] 
    \centering
    \caption{Quantitative comparison with state-of-the-art Gaussian Splatting fast optimization methods. ACE-GS consistently achieves superior rendering quality across all benchmarks while significantly reducing training duration compared to existing baselines. Results are marked as \capbest{best score}, \capsecond{second best score}, and \capthird{third best score}.}
    \label{tab:quantitative_results_simplified}
    
    \renewcommand{\arraystretch}{0.85} 
    \setlength{\extrarowheight}{1pt}   
    \setlength{\tabcolsep}{6pt} 
    \footnotesize
    
    \begin{tabular}{l ccc cc} 
        \toprule
        \textbf{Method} & \textbf{SSIM}$\uparrow$ & \textbf{PSNR}$\uparrow$ & \textbf{LPIPS}$\downarrow$ & \textbf{Time}$\downarrow$ & \textbf{FPS}$\uparrow$ \\
        \midrule
        \multicolumn{6}{c}{\textbf{Dataset: Mip-NeRF 360}} \\
        \midrule
        3DGS [11] & \second{0.813} & \third{27.54} & \second{0.221} & 24m 43s & 241.8 \\
        LightGaussian [5] & 0.801 & 27.03 & 0.245 & 36m 43s & 345.8 \\
        Compact3DGS [13] & 0.798 & 27.01 & 0.247 & 28m 06s & 331.0 \\
        Scaffold-GS [19] & \third{0.812} & \second{27.76} & \third{0.226} & \third{20m 57s} & \third{460.5} \\
        Speedy-Splat [9] & 0.782 & 26.89 & 0.295 & \second{14m 47s} & \best{1720.9} \\
        \textbf{ACE-GS(Ours)} & \best{0.821} & \best{28.10} & \best{0.215} & \best{5m 30s} & \second{745.6} \\

        \midrule
        \multicolumn{6}{c}{\textbf{Dataset: Tanks \& Temples}} \\
        \midrule
        3DGS [11] & \third{0.853} & \third{23.74} & \best{0.169} & 12m 50s & 313.0 \\
        LightGaussian [5] & 0.837 & 23.49 & 0.197 & 22m 30s & 580.7 \\
        Compact3DGS [13] & 0.833 & 23.36 & 0.200 & 15m 22s & 464.0 \\
        Scaffold-GS [19] & \second{0.854} & \second{24.13} & \third{0.174} & \third{11m 26s} & \third{704.9} \\
        Speedy-Splat [9] & 0.820 & 23.47 & 0.240 & \second{6m 48s} & \best{2216.0} \\
        \textbf{ACE-GS(Ours)} & \best{0.860} & \best{24.63} & \second{0.171} & \best{3m 01s} & \second{863.9} \\

        \midrule
        \multicolumn{6}{c}{\textbf{Dataset: Deep Blending}} \\
        \midrule
        3DGS [11] & \third{0.907} & 29.77 & \second{0.238} & 21m 58s & 268.0 \\
        LightGaussian [5] & 0.903 & 29.48 & 0.255 & 32m 34s & 442.9 \\
        Compact3DGS [13] & 0.905 & \third{29.87} & \third{0.254} & 22m 21s & 524.5 \\
        Scaffold-GS [19] & \second{0.908} & \second{30.20} & 0.261 & \third{14m 35s} & \third{828.5} \\
        Speedy-Splat [9] & 0.903 & 29.58 & 0.268 & \second{11m 11s} & \best{2393.0} \\
        \textbf{ACE-GS(Ours)} & \best{0.912} & \best{30.33} & \best{0.237} & \best{3m 00s} & \second{1108.4} \\
        \bottomrule
    \end{tabular}
\end{table}

\textbf{Implementation Details.} Our framework is implemented in PyTorch and built upon the 3DGS codebase. All experiments are conducted on a single NVIDIA RTX 4090 GPU with 24GB VRAM. For the geometric construction module, the densification interval $k_E$ is fixed at $100$ iterations. The sparsification threshold $\tau_{C}$ is set to $0.01$ to maintain a compact scene representation. For accuracy enhancement, the frequency scaling coefficient $\lambda_{A}$ is established as $1.0$ to achieve an optimal balance between high-frequency detail injection and numerical stability. All other hyper-parameters follow the default configurations of the original 3DGS.

\textbf{Datasets.} To comprehensively evaluate reconstruction performance across different capture protocols and environments, we utilize three representative benchmark datasets. For the Mip-NeRF 360 dataset~\cite{barron2022mip}, we employ nine scenes comprising five outdoor and four indoor environments with unbounded trajectories. Regarding the Tanks and Temples dataset~\cite{knapitsch2017tanks}, we adopt the Truck and Train scenes representing challenging large-scale outdoor environments. Furthermore, we utilize the Dr. Johnson and Playroom indoor scenes from the Deep Blending dataset~\cite{hedman2018deep} to analyze model behavior under unstructured capture conditions. All training and evaluation procedures are performed at the original image resolution.

\textbf{Baselines.} We evaluate ACE-GS against the original 3DGS~\cite{kerbl20233d} and incorporate representative methods across multiple dimensions, including Scaffold-GS~\cite{lu2024scaffold} for high-fidelity rendering and geometric accuracy, LightGaussian~\cite{fan2024lightgaussian} and Compact3DGS~\cite{lee2024compact} for compact scene representation, as well as the recent rapid training framework Speedy-Splat~\cite{hanson2025speedy} to verify our advantages in convergence acceleration and training efficiency.

\textbf{Metrics.} To assess reconstruction quality, we report three standard photometric metrics: PSNR, SSIM, and LPIPS. To evaluate training efficiency, we record the total wall-clock time from initialization to final convergence. For a fair comparison, the training duration for multi-stage methods such as LightGaussian is calculated as the cumulative sum of the pre-training phase and the subsequent refinement processes. Finally, we evaluate inference performance via FPS, where values are measured using CUDA events at the start and end of the forward rendering procedure.

\subsection{Results and Evaluation}

\begin{figure}[!t]
    \centering
    \includegraphics[width=\textwidth]{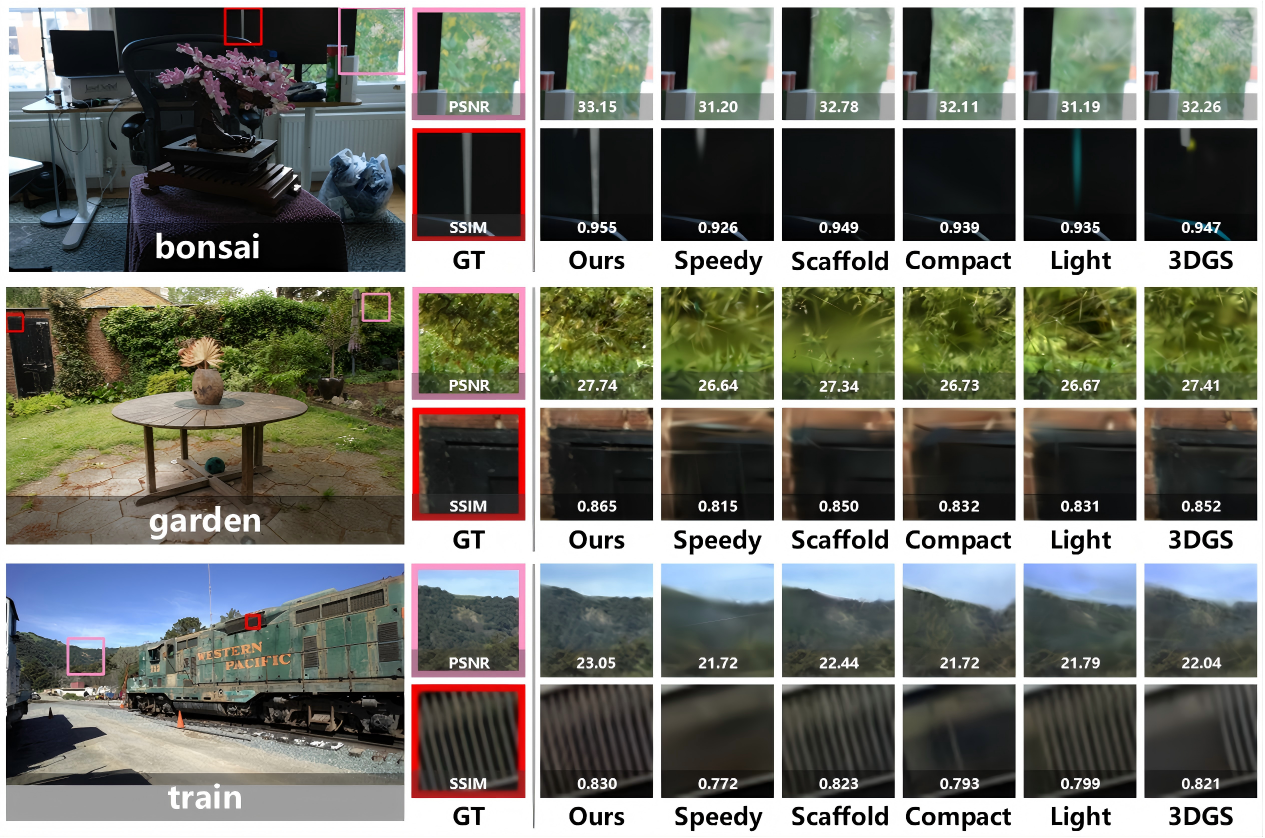}
    \caption{Visual comparison across diverse indoor and outdoor environments. Existing rapid optimization frameworks frequently suffer from severe geometric ambiguity or structural collapse in complex topological regions. In contrast, ACE-GS preserves accurate geometric silhouettes and sharp details through its intrinsic kinematic trajectory tracking, maintaining superior structural robustness where prior efficiency-oriented methods typically fail.}
    \label{fig:visual_multi_scene}
\end{figure}

\begin{figure}[!t]
    \centering
    \includegraphics[width=\textwidth]{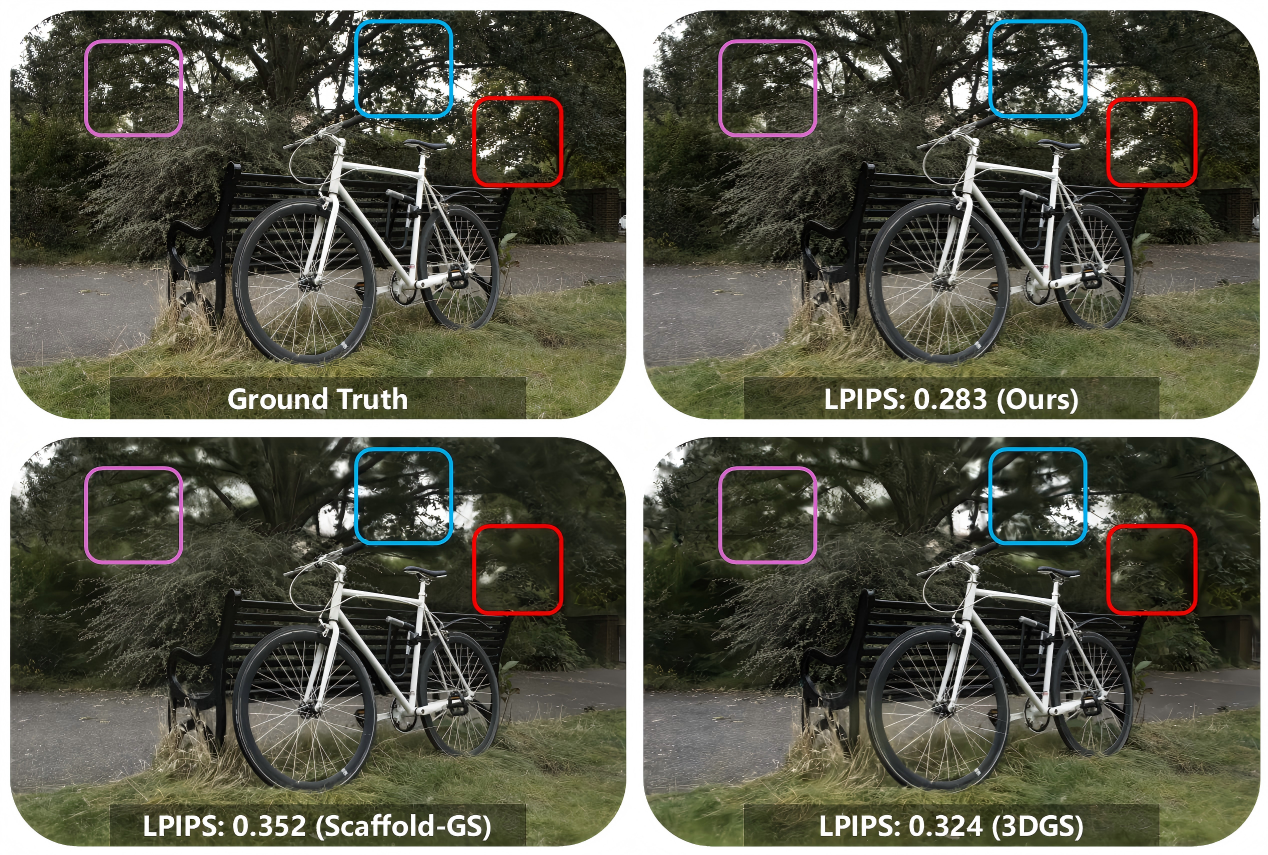}
    \caption{Qualitative results on the Bicycle scene highlighting high-frequency detail preservation. While representative baselines exhibit significant over-smoothing and loss of visual features in complex background regions, ACE-GS successfully restores intricate textures and fine structural boundaries by leveraging its unique cross-dimensional residual frequency compensation, achieving a significant leap in perceptual fidelity.}
    \label{fig:visual_bicycle_detail}
\end{figure}

\textbf{Quantitative Analysis.} As demonstrated in Table \ref{tab:quantitative_results_simplified}, ACE-GS exhibits significant performance superiority across all core quality metrics. Compared to the leading speed-oriented baseline, Speedy-Splat, ACE-GS achieves a 2.7$\times$ training speedup on the Mip-NeRF 360 dataset with a 1.21dB PSNR improvement. Against 3DGS, our method achieves an average speedup of over 4.5$\times$ while gaining 0.67dB in PSNR. This efficiency stems from the synergy between momentum consistency-guided densification and statistical sensitivity-driven online sparsification. By tracking primitive trajectories to ensure a topologically accurate geometric foundation and eliminating redundant primitives, ACE-GS overcomes the representation redundancy bottleneck. Furthermore, with the cross-dimensional residual frequency compensation, our method maintains ultra-fast training speeds without sacrificing reconstruction fidelity.

\textbf{Qualitative Analysis.} Visual comparisons in Fig. \ref{fig:visual_multi_scene} substantiate the dominant performance of ACE-GS in complex scene reconstruction. Existing rapid training frameworks frequently suffer from geometric ambiguity or structural collapse in complex topological regions, such as the dense metal grilles in the Train scene. In contrast, ACE-GS restores exceptionally clear silhouettes and sharp high-frequency details. This is achieved by identifying the true physical manifold from stochastic gradients and pruning redundant primitives with negligible geometric contributions. 

The reconstruction efficacy for ultra-fine textures is further validated in Fig. \ref{fig:visual_bicycle_detail}. Taking the Bicycle scene as an example, prior efficiency-oriented paradigms often result in over-smoothing in high-frequency regions (e.g., distant vegetation) to accelerate convergence. Conversely, ACE-GS re-injects high-order texture energy into primitive attributes via our frequency compensation mechanism. This enables limited primitives to capture rich visual information with higher representation efficiency, maintaining extraordinary perceptual fidelity even in deep and complex backgrounds.

\subsection{Ablation Studies}
\label{sec:ablation}

\begin{table}[t]
\centering
\caption{Quantitative ablation study on the Bicycle scene. The results demonstrate that ACE-GS achieves an optimal balance between reconstruction quality and efficiency through the synergy of its core modules. Removing any component disrupts this equilibrium, leading to a noticeable drop in fidelity or convergence speed. This confirms that each module is essential for maintaining high-performance modeling under rapid optimization.}
\label{tab:ablation}
\begin{tabular}{lccccc}
\toprule
 & \footnotesize SSIM $\uparrow$ & \footnotesize PSNR $\uparrow$ & \footnotesize LPIPS $\downarrow$ & \footnotesize Time $\downarrow$ & \footnotesize FPS $\uparrow$ \\
\midrule
w/o A & 0.743 & 25.08 & 0.246 & 5m 36s & 783.2 \\
w/o C & 0.757 & 25.24 & 0.245 & 5m 52s & 648.2 \\
w/o E & 0.750 & 25.19 & 0.242 & 10m 04s & 494.8 \\
\midrule
\textbf{ACE-GS} & \textbf{0.753} & \textbf{25.47} & \textbf{0.243} & \textbf{5m 44s} & \textbf{732.2} \\
\bottomrule
\end{tabular}
\end{table}

\begin{figure}[t]
\centering
\includegraphics[width=\linewidth]{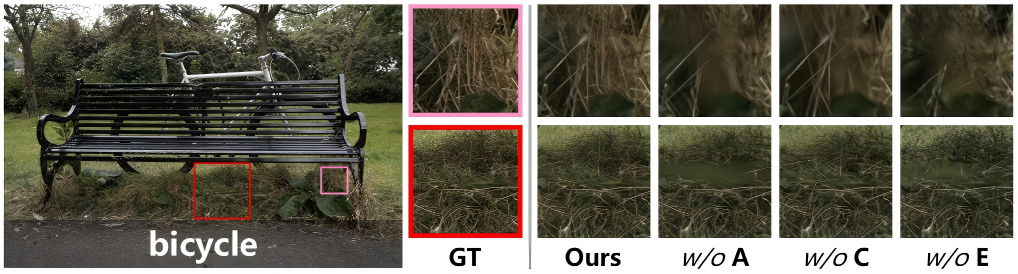}
\caption{Ablation study of proposed components on the Bicycle scene. Removing our proposed modules leads to visible degradation in the rendered results, especially in fine-structured regions such as the grass and spokes. Specifically, omitting frequency compensation A causes significant smoothing of high-frequency textures, while disabling online sparsification C results in blurred structural boundaries due to redundant primitive accumulation. Furthermore, removing momentum-guided densification E introduces severe floating artifacts. ACE-GS produces high-quality results even in these challenging areas.}
\label{fig:ablation_visual}
\end{figure}

\noindent We evaluate the individual contribution of each component on the \textit{Bicycle} scene, with quantitative and qualitative results summarized in Table \ref{tab:ablation} and Fig. \ref{fig:ablation_visual}, respectively. The results demonstrate that removing any module disrupts the equilibrium between reconstruction fidelity and efficiency. Specifically, omitting residual frequency compensation (A) causes a 0.39dB drop in PSNR and results in over-smoothed high-frequency textures, particularly on grass and spokes, confirming the necessity of high-order energy injection for detail recovery. Disabling online sparsification (C) leads to a noticeable decrease in rendering rate and blurred structural boundaries due to the accumulation of redundant primitives, highlighting its role in maintaining a lightweight representation. Most critically, removing momentum-guided densification (E) increases training time by approximately 1.75$\times$ and introduces severe floating artifacts. This validates that momentum consistency is essential for regularizing the optimization path and ensuring a clean geometric manifold. In contrast, the full ACE-GS model achieves superior reconstruction quality while ensuring the best overall optimization and rendering performance.

\subsection{Hyper-parameter Sensitivity Analysis}
\label{sec:sensitivity}

\begin{figure*}[t]
    \centering
    \includegraphics[width=1.0\linewidth]{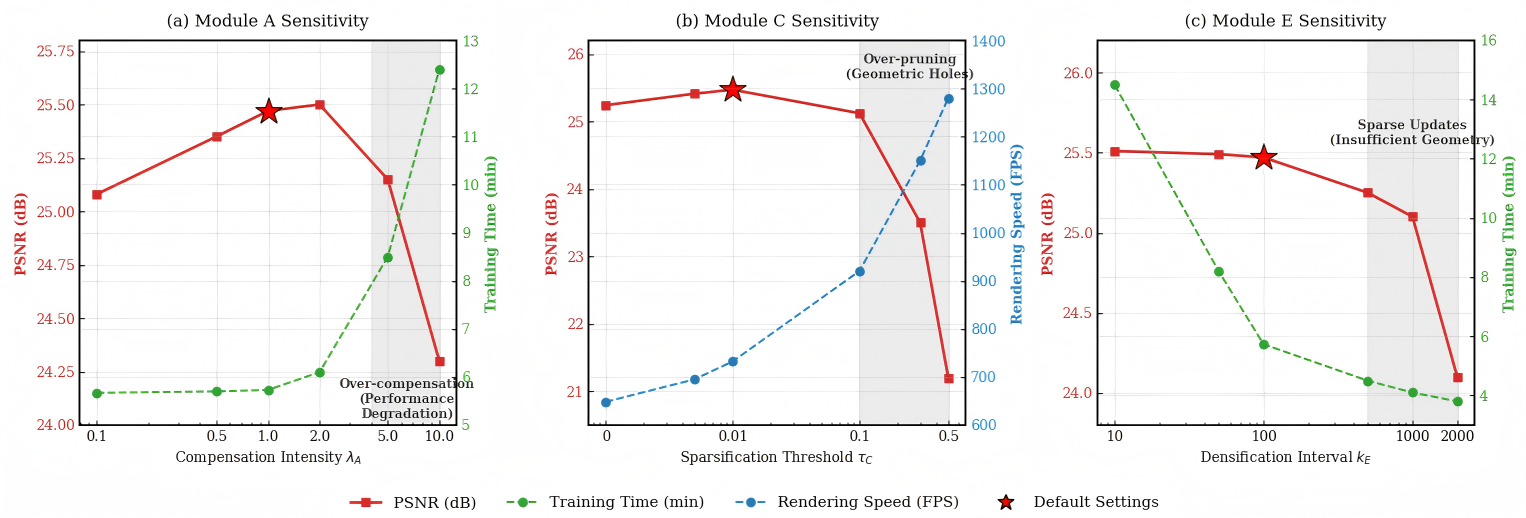}
    \caption{\textbf{Hyper-parameter sensitivity analysis of the proposed ACE-GS.} We evaluate the impact of compensation intensity $\lambda_A$ in (a), sparsification threshold $\tau_C$ in (b), and densification interval $k_E$ in (c) on visual quality measured by PSNR and computational efficiency measured by training time or FPS. Red stars indicate our default settings, while shaded areas represent undesirable regions characterized by significant performance drops, excessive computational costs, or geometric artifacts.}
    \label{fig:hyper_analysis}
\end{figure*}

\textbf{Hyper-parameter Sensitivity Analysis.} To evaluate the robustness of ACE-GS, we analyze the impact of key hyper-parameters as illustrated in Fig.~\ref{fig:hyper_analysis}. For the compensation intensity $\lambda_A$, values near 1.0 yield an optimal quality-efficiency trade-off; excessive intensity introduces high-frequency artifacts and unnecessary computational overhead. Regarding the sparsification threshold $\tau_C$, although higher values significantly boost rendering FPS, they eventually cause a sharp decline in PSNR due to over-pruning. Finally, the densification interval $k_E$ balances geometric completeness with training speed, where intervals that are too long lead to insufficient representation. Based on these observations, we set $\lambda_A=1.0$, $\tau_C=0.01$, and $k_E=100$ as the default configurations to maintain stable performance across diverse scenes.

\section{Limitations and Future Works}
\label{sec:limitation}

\begin{table*}[t]
\centering
\caption{\textbf{Evaluation of resource overhead.} Bold font denotes the most compact baseline methods with the lowest values. Compared with specialized compression frameworks, ACE-GS shows significant insufficiencies in memory consumption and primitive counts.}
\vspace{-2mm}
\label{tab:resource_limitation}
\resizebox{\linewidth}{!}{
\begin{tabular}{l|cc|cc|cc}
\toprule
\multirow{2}{*}{\textbf{Method}} & \multicolumn{2}{c|}{\textbf{Mip-NeRF 360}} & \multicolumn{2}{c|}{\textbf{Tanks \& Temples}} & \multicolumn{2}{c}{\textbf{Deep Blending}} \\ \cline{2-7} 
 & Mem (MB) $\downarrow$ & Gauss (M) $\downarrow$ & Mem (MB) $\downarrow$ & Gauss (M) $\downarrow$ & Mem (MB) $\downarrow$ & Gauss (M) $\downarrow$ \\ \midrule
Gaussian-Splatting & 623.65 & 2.637 & 372.40 & 1.575 & 586.34 & 2.479 \\
LightGaussian & \textbf{41.06} & 0.899 & \textbf{24.71} & 0.536 & \textbf{38.23} & 0.840 \\
Compact 3DGS & 46.87 & 1.405 & 37.65 & 0.836 & 41.04 & 1.045 \\
Speedy-Splat & 70.09 & \textbf{0.296} & 43.11 & \textbf{0.182} & 59.41 & \textbf{0.251} \\ \midrule
\rowcolor[HTML]{EFEFEF} 
\textbf{ACE-GS (Ours)} & 277.13 & 1.172 & 127.96 & 0.533 & 153.36 & 0.649 \\ \bottomrule
\end{tabular}
}
\vspace{-3mm}
\end{table*}

\noindent While ACE-GS accelerates training by $2.78\times$ / $3.70\times$ / $3.90\times$ on Mip-NeRF 360, Tanks \& Temples, and Deep Blending respectively, its resource efficiency remains limited as shown in Table \ref{tab:resource_limitation}. Prioritizing geometric fidelity results in denser representations and higher memory usage than models optimized for extreme sparsity. Additionally, rendering throughput trails behind specialized inference frameworks, restricting practicality in hardware-constrained scenarios. Future research will explore quantization and adaptive pruning synergy to develop efficient architectures that harmonize quality and storage for improved suitability on resource-constrained platforms.

\section{Conclusion}
In this paper, we presented ACE-GS, a framework designed to balance accuracy, compactness, and efficiency in 3D Gaussian Splatting. By treating reconstruction as a dynamic primitive allocation problem, we effectively addressed the over-smoothing and slow convergence issues typical of accelerated paradigms. Our approach integrates momentum-guided densification, sensitivity-driven sparsification, and residual frequency compensation to ensure robust geometry and fine-grained detail restoration. Extensive experiments demonstrate that ACE-GS achieves significant training acceleration while maintaining superior visual fidelity and structural integrity over competitive baselines. This system offers a practical and effective solution for the rapid representation of complex 3D scenes in just a few minutes.

%
%
\bibliographystyle{splncs04}
\bibliography{main}
\end{document}